\documentclass[letterpaper]{article} 
\usepackage{aaai2026}  
\usepackage{times}  
\usepackage{helvet}  
\usepackage{courier}  
\usepackage[hyphens]{url}  
\usepackage{graphicx} 
\usepackage{natbib}  
\usepackage{caption} 
\usepackage{algorithm}
\usepackage{algorithmic}

\usepackage{amssymb}
\usepackage{amsmath}
\usepackage{booktabs}
\usepackage{multirow}
\usepackage{array}
\usepackage[table]{xcolor} 
\usepackage{makecell}  
\usepackage{colortbl}       
\usepackage{comment}
\nocopyright

\usepackage{newfloat}
\usepackage{listings}
\DeclareCaptionStyle{ruled}{labelfont=normalfont,labelsep=colon,strut=off} 
\floatstyle{ruled}
\newfloat{listing}{tb}{lst}{}
\floatname{listing}{Listing}
\title{WaveFormer: Frequency-Time Decoupled Vision Modeling with Wave Equation}
\author{
    Zishan Shu\textsuperscript{\rm 1}\equalcontrib,
    Juntong Wu\textsuperscript{\rm 1}\equalcontrib,
    Wei Yan\textsuperscript{\rm 1}\equalcontrib,
    Xudong Liu\textsuperscript{\rm 1},
    Hongyu Zhang\textsuperscript{\rm 1},
    Chang Liu\textsuperscript{\rm 2}\thanks{Corresponding authors.},\\
    Youdong Mao\textsuperscript{\rm 1}$^\dagger$
,
    Jie Chen\textsuperscript{\rm 1}$^\dagger$
}

\affiliations {
    \textsuperscript{\rm 1}Peking University\\
    \textsuperscript{\rm 2}Tsinghua University\\
    \{zishanshu, wujt, weiyan, xudongliu, zhanghy\}@stu.pku.edu.cn\\
    \{ymao, jiechen2019\}@pku.edu.cn\\
    liuchang2022@tsinghua.edu.cn
}

\usepackage{bibentry}

\begin{document}

\maketitle

\begin{abstract}
Vision modeling has advanced rapidly with Transformers, whose attention mechanisms capture visual dependencies but lack a principled account of how semantic information propagates spatially. We revisit this problem from a wave-based perspective: feature maps are treated as spatial signals whose evolution over an internal propagation time (aligned with network depth) is governed by an underdamped wave equation. In this formulation, spatial frequency—from low-frequency global layout to high-frequency edges and textures—is modeled explicitly, and its interaction with propagation time is controlled rather than implicitly fixed. We derive a closed-form, frequency–time decoupled solution and implement it as the Wave Propagation Operator (WPO), a lightweight module that models global interactions in $\mathcal{O}(N \log N)$ time—far lower than attention. Building on WPO, we propose a family of WaveFormer models as drop-in replacements for standard ViTs and CNNs, achieving competitive accuracy across image classification, object detection, and semantic segmentation, while delivering up to $1.6\times$ higher throughput and 30\% fewer FLOPs than attention-based alternatives. Furthermore, our results demonstrate that wave propagation introduces a complementary modeling bias to heat-based methods, effectively capturing both global coherence and high-frequency details essential for rich visual semantics. Codes are available at:
https://github.com/ZishanShu/WaveFormer.
\end{abstract}


\section{Introduction}
\begin{figure}[t]
    \centering
    \vspace{-3pt}
    \includegraphics[width=\linewidth]{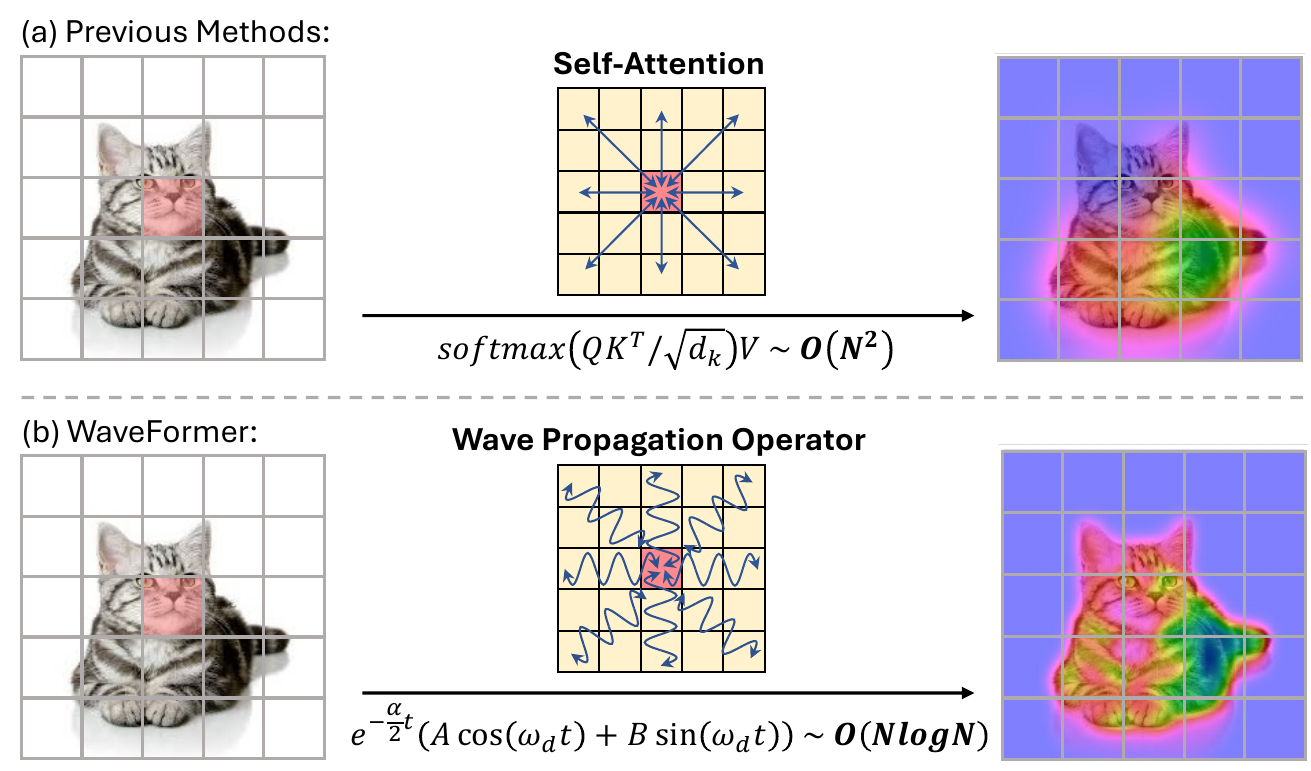} 
    \caption{(a) The self-attention operator facilitates information from a pixel to all other pixels, resulting in $\mathcal{O}(N^2)$ complexity.(b) The wave propagation operator (WPO) introduces oscillatory dynamics that balance energy across different frequency components, enabling new modeling behavior and reduced complexity.}
    \label{fig:pic/fig2}
    \vspace{-3ex}
\end{figure}
Deep learning has achieved remarkable success in visual representation learning, largely driven by convolutional neural networks (CNNs)\cite{he2016deep, krizhevsky2012imagenet} and vision transformers (ViTs)\cite{dosovitskiy2021image, liu2021swin}. These architectures model local patterns and long-range dependencies via similarity-based interactions between tokens, but they operate mainly in the spatial domain and lack an explicit inductive bias on how information at different spatial frequencies should be propagated, which can hinder the joint preservation of global semantics and fine-grained details~\cite{luo2016understanding}.

A complementary line of work views global feature propagation through the lens of partial differential equations (PDEs). Throughout this paper, we use frequency to refer to the spatial frequency of 2D feature maps (separating low-frequency layout from high-frequency edges and textures), and time to denote the continuous propagation variable of the PDE, corresponding in networks to the temporal evolution of features. Most existing physics-inspired designs adopt heat-like conduction, which in the Fourier domain acts as a strong low-pass filter: the decay of each mode grows with the product of its spatial frequency and propagation time. This tight frequency–time coupling causes high-frequency components to vanish much faster than low-frequency ones, leading to over-smoothed features and weak modeling of local structures.

We therefore ask: how should we propagate information across space so that the interaction between spatial frequency and propagation time is explicitly controllable, rather than intrinsically low-pass? To address this, we adopt a wave-based formulation. In contrast to heat-based conduction, the underdamped wave equation governs oscillatory evolution of spatial signals, where amplitude decay is controlled by damping and is not intrinsically tied to spatial frequency. In the frequency domain, low- and high-frequency components can coexist and exchange information over propagation time, providing a physically grounded alternative to similarity-based attention and heat-based conduction that preserves fine details while still supporting long-range transport.

We formalize this idea by deriving the general solution of the underdamped wave equation in 2D and extending it to high-dimensional feature space. Building upon this formulation, we propose the Wave Propagation Operator (WPO), a novel module that enables frequency-aware semantic modeling via global oscillatory propagation. Moreover, it achieves \(\mathcal{O}(N \log N)\) complexityand captures multi-scale interactions in a frequency–time decoupled fashion. This design introduces a novel physics-inspired inductive bias by leveraging the underdamped wave equation to enable oscillatory information propagation. This mechanism captures multi-scale semantic features while preserving structural and fine-grained details, providing a frequency-aware and physically grounded solution for visual representation learning, as illustrated in Fig.~\ref{fig:pic/fig2}. We further develop a family of WaveFormer models (i.e., WaveFormer-Tiny/Small/Base) and conduct extensive experiments to demonstrate their effectiveness across a wide range of visual tasks. Compared to benchmark vision backbones with diverse architectures (e.g., ConvNeXt~\cite{liu2022convnet}, Swin Transformer~\cite{liu2021swin}, Vision Mamba~\cite{zhu2024visionmamba}, and vHeat~\cite{Wang_2025_CVPR}), WaveFormer consistently achieves superior performance on image classification, object detection, and semantic segmentation across different model scales. Specifically, WaveFormer-Base achieves 84.2\% Top-1 accuracy on ImageNet-1K, surpassing Swin Transformer by 0.7\%. 

Our main contributions can be summarized as follows: 
\begin{itemize} 
\item We propose a novel \textbf{frequency–time decoupled modeling framework} for visual semantic propagation, which addresses the limitations of heat-based methods. To implement this framework, we adopt damped wave dynamics as the underlying mechanism, enabling oscillatory propagation that preserves both global semantics and fine-grained details across time. 
\item We instantiate this modeling principle through \textbf{WaveFormer}, a physics-inspired vision backbone built upon the Wave Propagation Operator (WPO), which simulates damped wave propagation in the frequency domain. WPO achieves efficient and interpretable global information transfer with $\mathcal{O}(N \log N)$ complexity. 
\item We develop WaveFormer-Tiny/Small/Base and demonstrate their effectiveness across image classification, object detection, and semantic segmentation, achieving \textbf{state-of-the-art accuracy–efficiency trade-offs} in terms of throughput, FLOPs, and performance. 
\end{itemize}

\section{Related Work}
\subsection{Vision Foundational Models}
Vision Foundational Models have evolved significantly from early convolutional architectures to modern large-scale pre-trained backbones. Convolutional Neural Networks (CNNs)~\cite{he2016deep, krizhevsky2012imagenet} marked the initial breakthrough in visual perception by leveraging local convolution kernels for spatially invariant feature extraction. With the advent of large-scale datasets~\cite{deng2009imagenet} and high-performance GPUs, increasingly deeper and more efficient architectures~\cite{he2016deep, huang2017densely, simonyan2014very, szegedy2015going, howard2017mobilenets, radosavovic2020designing, tan2019efficientnet, yang2021focal} have been proposed. To address the inherent limitation of local receptive fields, various modifications have been introduced, such as dilated convolutions~\cite{chollet2017xception}, large-kernel convolutions~\cite{ding2022scaling}, and dynamic convolutions~\cite{dai2017deformable, wang2023internimage}. However, despite these advances, CNNs still struggle to efficiently capture long-range dependencies in high-resolution images.

Vision Transformers (ViTs)~\cite{dosovitskiy2021image} further revolutionized vision modeling by introducing self-attention, which inherently models global dependencies between all image patches. Pretrained on massive datasets~\cite{dosovitskiy2021image, peng2022beitv2, touvron2021deit}, ViTs have become dominant in various vision tasks. Hierarchical variants~\cite{ding2022davit, dong2022cswin, liu2021swin, tian2023iptpn, zhang2023hivit, zhao2022graformer} improved scalability and representation power, but their quadratic $\mathcal{O}(N^2)$ complexity remains a bottleneck for high-resolution images. Numerous approaches have been proposed to reduce the computational cost, including windowed attention, linear attention, and cross-covariance attention~\cite{ali2021xcit, chen2021crossvit, liu2021swin}, but often at the expense of receptive field or non-linear modeling capacity. Hybrid architectures combining convolutional and transformer modules were also explored to balance efficiency and performance.

More recently, State Space Models (SSMs)~\cite{gu2022parameterization, nguyen2022s4nd, wang2023selective} have emerged as an alternative, offering long-sequence modeling with linear complexity, which also migrated from the natural language area (Mamba~\cite{gu2023mamba}). Adaptations like visual SSMs~\cite{liu2024vmamba, zhu2024vision} introduced selective scanning for 2D images, while RWKV and RetNet~\cite{peng2023rwkv, sun2023retentive} improved parallelism by combining transformer-style training with efficient RNN-style inference. Nevertheless, modeling a 2D image as a 1D sequence often impairs spatial interpretability.

\begin{figure*}[ht]
    \centering
    \includegraphics[width=\linewidth]{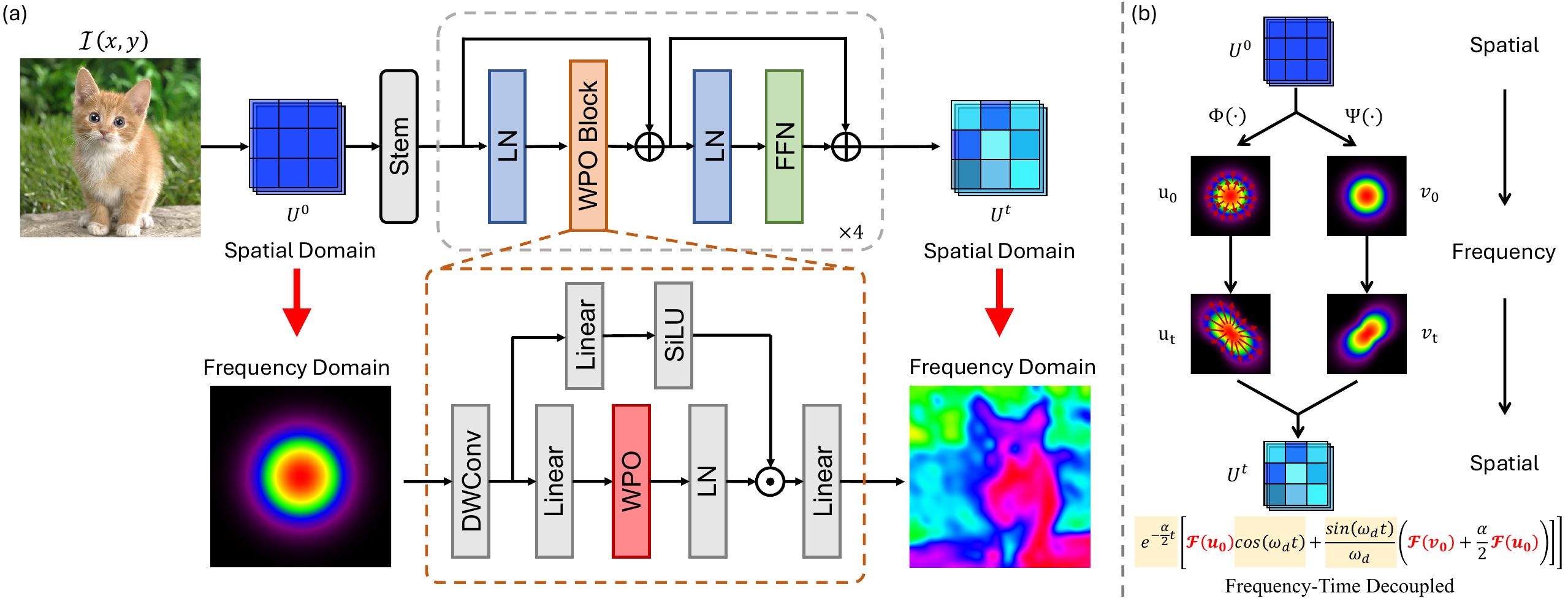} 
    \caption{The network architecture of WaveFormer. (a) The network follows a hierarchical vision backbone design, consisting of a stem followed by four stages of WPO Blocks, where each stage integrates the Wave Propagation Operator (WPO) with feed-forward layers and downsampling in between. (b) Implementation of the WPO. The input feature map is transformed into the frequency domain, where the frequency–time decoupled analytical solution of the underdamped wave equation modulates each frequency component through oscillatory dynamics. The result is then mapped back to the spatial domain via the inverse Fourier transform, enabling global semantic propagation while preserving fine-grained high-frequency details.}
    \label{fig:pic/fig3}
    \vspace{-2ex}
\end{figure*}

\subsection{Biology and Physics Inspired Models}
Biology and physics have long inspired the design of visual models, providing complementary inductive biases. Diffusion models~\cite{ho2020ddpm, saharia2022imagen, song2020ddim}, grounded in non-equilibrium thermodynamics~\cite{degroot2013nonequilibrium}, model image generation as a Markovian denoising process. QB-Heat~\cite{chen2022heatssl} utilizes the physical heat equation as a supervision signal for masked image modeling, mimicking thermal diffusion to propagate semantic information. Similarly, Spiking Neural Networks (SNNs)~\cite{ghosh2009snn, lee2016snnbackprop, tavanaei2019snnreview}, inspired by biological neurons, simulate discrete spike-driven information transmission and have been applied to low-level vision tasks~\cite{bawane2018snnocr}.

Beyond biological analogies, physical principles of signal propagation have also been explored. Heat conduction-based, such as vHeat~\cite{Wang_2025_CVPR},  approaches inherently favor smooth, monotonic diffusion but tend to oversmooth high-frequency components, losing fine-grained details. Prior studies on wave propagation~\cite{sorteberg2018approximatingsolutionwavepropagation} demonstrate the potential of oscillatory dynamics for modeling frequency-balanced signal evolution~\cite{Rasht_Behesht_2022}. Such models preserve both global semantics and local textures, providing a physics-consistent alternative to purely data-driven architectures.

\section{Methodology}
\begin{figure*}[ht]
    \centering
    \includegraphics[width=\linewidth]{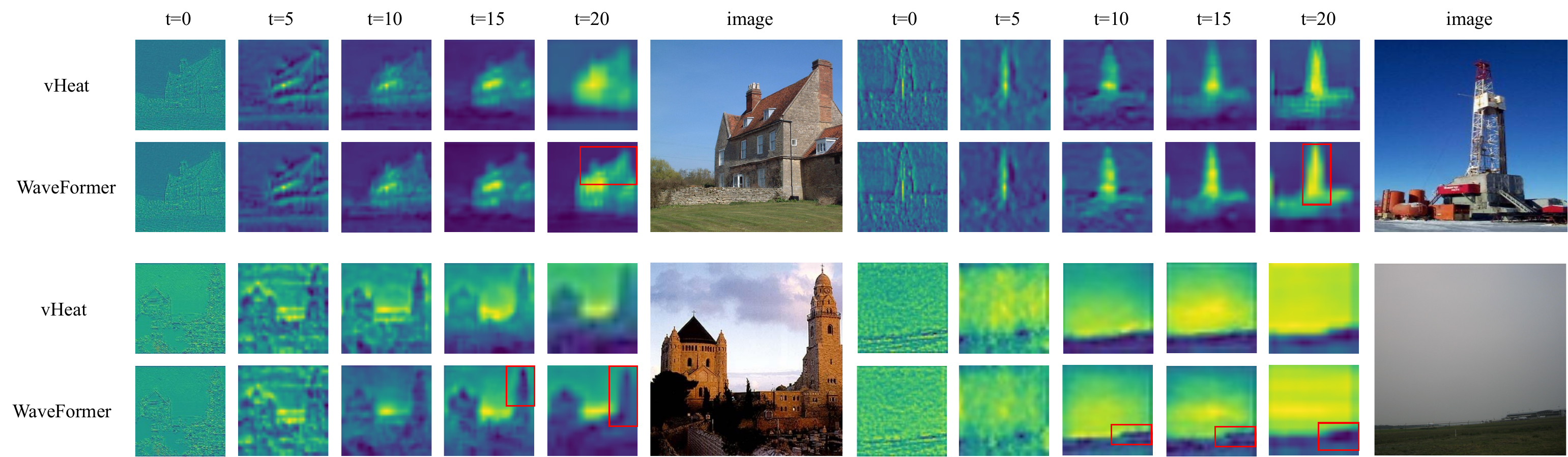} 
    \caption{Attention map evolution over time for heat conduction (top) and wave propagation (bottom) across different ADE20K image cases. Red boxes highlight key different regions.}
    \label{fig:pic/case}
\end{figure*}

This chapter presents the wave physics-inspired approach for visual modeling, formulating feature evolution as semantic propagation via a damped wave equation. Building on this foundation, we introduce WaveFormer, a novel vision backbone that incorporates the newly proposed Wave Propagation Operators (WPO) to achieve frequency-time decoupled semantic transfer, enabling preservation of both global structure and local details.

\vspace{-5pt}
\subsection{Semantic Propagation Formalization}

We formalize semantic propagation as damped wave dynamics, where visual features evolve through oscillatory patterns over space and time. Step 1 defines the initial semantic and velocity fields; Step 2 decomposes the solution into spatial modes with frequency–time decoupled dynamics; Step 3 derives a closed-form frequency-domain formulation for efficient and interpretable propagation. This wave-based approach preserves both global structure and fine-grained detail, offering a principled alternative to heat-based modeling.


\paragraph{Step 1. Wave Equation-based Semantic Propagation}
Denote an input image $\mathcal{I}(x,y)$ over a bounded spatial domain as $(x, y) \in [0, H] \times [0, W]$. For semantic propagation, we initialize its semantic field $u(x,y,t)$) and optional semantic velocity field $v(x,y,t)$ as
\begin{equation}
\left\{
\begin{aligned}
u(x,y,0) &= \Phi(\mathcal{I}(x,y)) &= u_0, \\
v(x,y,0) &= \Psi(\mathcal{I}(x,y)) &= v_0,
\end{aligned}
\right.
\label{1}
\end{equation}
where $\Phi(\cdot)$ is a semantic encoder that maps the image to an initial activation field $u_0$, and $\Psi(\cdot)$ optionally provides a velocity field $v_0$ that reflects the rate of semantic excitation or suppression across regions. 

The evolution of the semantic field $u(x,y,t)$ can be modeled by the damped wave equation, as
\begin{equation}
\begin{split}
\frac{\partial^2 u}{\partial t^2} + \alpha \frac{\partial u}{\partial t}
&= v^2 \left( \frac{\partial^2 u}{\partial x^2} + \frac{\partial^2 u}{\partial y^2} \right), \\
u(x,y,0) &= u_0,\quad \left.\frac{\partial u}{\partial t}\right|_{t=0} = v_0.
\end{split}
\label{3}
\end{equation}
This equation governs the propagation of semantic signals, where $v$ controls the propagation velocity, and $\alpha$ determines the damping behavior across time.

\paragraph{Step 2. Frequency–Time Decoupling with Oscillatory Mode Decomposition}

To analyze the multi-scale nature of semantic propagation, we decompose the wave field $u(x, y, t)$ into space-time separable modes. We assume a spatially orthogonal sine basis $\phi_{n,m}(x,y) = \sin\left( \frac{n \pi x}{H} \right)\sin\left( \frac{m \pi y}{W} \right)$, where $(n,m)$ indexes horizontal and vertical frequency components. The field can thus be expressed as:
\vspace{-3ex}
\begin{equation}
\begin{aligned}
u(x,y,t) &= \sum_{n=1}^\infty \sum_{m=1}^\infty q_{n,m}(t) \cdot \phi_{n,m}(x,y) \\
         &= \sum_{n=1}^\infty \sum_{m=1}^\infty 
         e^{-\frac{\alpha}{2}t} \left[ A_{n,m} \cos(\omega_d t) + B_{n,m} \sin(\omega_d t) \right] \\
         &\quad \times \sin\left( \frac{n \pi x}{H} \right) \sin\left( \frac{m \pi y}{W} \right),
\end{aligned}
\label{6}
\end{equation}
where the constants $A_{n,m}$ and $B_{n,m}$ are determined by the initial semantic field $u_0(x,y)$ and velocity field $v_0(x,y)$ via Fourier projection. 

Unlike arbitrary learnable functions, this structure is not free-form but derived from the physical solution of wave dynamics. Notably, $q_{n,m}(t)$ reveals a frequency-time decoupled structure: the damping term $e^{-\frac{\alpha}{2}t}$ uniformly governs temporal decay across all modes, while the oscillation frequency $\omega_d$ is determined solely by the spatial mode index $(n,m)$ and wave velocity $v$. The resulting damped frequency is:
\begin{equation}
\omega_{n,m} = v \sqrt{\left( \frac{n\pi}{H} \right)^2 + \left( \frac{m\pi}{W} \right)^2}, \quad
\omega_d = \sqrt{\omega_{n,m}^2 - \left( \frac{\alpha}{2} \right)^2}.
\label{6a}
\end{equation}

This formulation enables frequency–aware semantic propagation: low-frequency modes govern global structure, and high-frequency modes preserve local detail. The temporal and spectral behaviors are independently controllable—an essential property for modeling long-range, multi-scale semantic interaction without over-smoothing.

\paragraph{Step 3. Closed-Form Solution for Practice}

To facilitate practical implementation, we derive a closed-form solution to the damped wave equation (Eq.~\eqref{3}) in the frequency domain. Applying the 2D Fourier transform $\mathcal{F}$ to both sides of the PDE, we convert spatial derivatives into multiplicative frequency terms and obtain the following explicit solution in the spectral domain:
\begin{equation}
\begin{aligned}
u(x,y,t) = \mathcal{F}^{-1} &\Big\{ e^{-\frac{\alpha}{2} t} \Big[ \mathcal{F}(u_0) \cos(\omega_d t) \\
&+ \frac{\sin(\omega_d t)}{\omega_d} \Big( \mathcal{F}(v_0) + \frac{\alpha}{2} \mathcal{F}(u_0) \Big) \Big] \Big\},
\end{aligned}
\label{7}
\end{equation}
where $u_0$ and $v_0$ denote the initial semantic field and velocity as defined in Eqs.~\eqref{1}, and $\omega_d$ is the damped propagation frequency defined in Eq.~\eqref{6a}.

This solution serves as a continuous counterpart to the mode expansion in Eq.~\eqref{6}, shifting from discrete spatial modes to continuous frequency components. Notably, it reveals a desirable frequency–time decoupling: unlike the heat equation where high frequencies decay quickly due to the $e^{-k \omega^2 t}$ term, the damping $e^{-\alpha t / 2}$ applies uniformly across frequencies, while $\omega_d$ governs frequency-specific oscillations. This structure allows both low- and high-frequency semantics to persist over time, rather than being rapidly suppressed. The damping behavior remains controllable via $\alpha$, enabling fine-grained temporal regulation.

  
  
  

Together, our wave-based formulation models semantic propagation as damped oscillations over space and time. By decoupling frequency and time, it preserves both global coherence and fine details without over-smoothing. This provides a compact, physically grounded foundation for expressive semantic modeling in WaveFormer.

\subsection{Semantic Propagation Instantiation}
\paragraph{Wave Propagation Operator (WPO)}
Drawing inspiration from the analogy between wave physics and the propagation of visual semantics, we propose the Wave Propagation Operator (WPO), a frequency-modulated propagation module designed to simulate wave-based semantic transfer within image representations.

WPO is grounded in the oscillatory dynamics of the damped wave equation, as formalized in Eq.~\eqref{7}. The solution exhibits temporally coherent oscillation, frequency-independent damping, and preserves both global and fine-grained semantic components over time.

To implement WPO in practice, we extend the 2D scalar semantic field $u(x,y,t)$ to a multi-channel semantic tensor $U(x,y,c,t)$, where $c = 1, \ldots, C$ is the channel index in a feature map. The input $U^0 = U(x,y,c,0)$ and its optional initial velocity $V^0 = V(x,y,c,0)$ are propagated across abstract time $t$ using the frequency-domain closed-form solution. Formally, WPO) can be defined as:
\begin{equation}
\begin{aligned}
U^t = \mathcal{F}^{-1} &\Big\{ e^{-\frac{\alpha}{2} t} \Big[ \mathcal{F}(U^0)\cos(\omega_d t) \\
&\quad + \frac{\sin(\omega_d t)}{\omega_d} \Big( \mathcal{F}(V^0) + \frac{\alpha}{2}\,\mathcal{F}(U^0) \Big) \Big] \Big\},
\end{aligned}
\label{8}
\end{equation}
where $\mathcal{F}$ and $\mathcal{F}^{-1}$ denote the 2D Fourier transform and its inverse applied per channel, and $\omega_d$ is the damped frequency magnitude computed from the spatial frequency spectrum:
\begin{equation}
\omega_d = \sqrt{v^2 (\omega_x^2 + \omega_y^2) - \left( \tfrac{\alpha}{2} \right)^2}.
\label{9}
\end{equation}

Here, $(\omega_x, \omega_y)$ denotes the 2D frequency grid over spatial dimensions. The scalar hyperparameters $\alpha$ and $v$ control the damping and semantic propagation velocity, respectively, and can be made trainable or adaptive.

\paragraph{WaveFormer}
Inspired by the oscillatory nature of physical wave propagation, we propose WaveFormer, a physics-inspired architecture for visual representation learning. Unlike heat conduction, which irreversibly dissipate information, WaveFormer models semantic propagation as damped oscillations, enabling global context modeling with reversible energy transfer.

WaveFormer serves as a drop-in replacement for standard ViTs or CNNs. As illustrated in Fig.~\ref{fig:pic/fig3}, the model partitions the input image into patches and embeds them into a feature map. Multiple stages progressively downsample the spatial resolution while increasing the channel dimension.

Each stage consists of several Wave Propagation Layers, where Transformer self-attention is replaced by WPO, while depthwise convolutions and feedforward layers enhance local and channel-wise modeling. WaveFormer is built upon the Wave Propagation Operator (WPO), which integrates the principle of the damped wave equation (Eq.~\ref{8}) into discrete feature processing. The wave velocity and damping are made adaptive to image content, making WaveFormer a flexible and learnable vision backbone.

We develop three variants: WaveFormer-Tiny, WaveFormer-Small, and WaveFormer-Base, differing in width and depth. Thanks to its physically grounded propagation dynamics and efficient spectral implementation, WaveFormer achieves competitive accuracy with reduced computational cost compared to attention-based backbones.

\subsection{Discussion}

\paragraph{Why does the damped wave field $U(x, y, c, t)$ correspond to visual features, and what is the role of the damped oscillation terms $\cos(\omega_d t)$ and $\sin(\omega_d t)$ in WaveFormer?}  
WaveFormer models semantic information as a damped wave field evolving over time. The feature value $U(x, y, c, t)$ captures oscillatory information flow across the feature map, enabling bidirectional, reversible propagation—unlike heat conduction, which diffuses irreversibly. In the frequency domain, each spatial frequency $(\omega_x, \omega_y)$ propagates via a damped oscillation:  
\[
e^{-\frac{\alpha}{2} t} \left[ 
\hat{u}_0 \cos(\omega_d t) 
+ \frac{1}{\omega_d} \left( \hat{v}_0 + \frac{\alpha}{2} \hat{u}_0 \right) \sin(\omega_d t)
\right],
\]
where $\omega_d$ is the damped frequency determined by spatial frequency $\omega_0$ and learnable damping $\alpha$. Low frequencies decay slowly and encode global structure, while high frequencies oscillate rapidly and are selectively attenuated, preserving edges and textures. This frequency-aware mechanism allows WaveFormer to retain both global context and fine detail, enabling interpretable semantic propagation.

\paragraph{Why is wave propagation suitable for visual representation learning?}  
Many physics-inspired methods use heat conduction for global modeling, but this oversmooths features and suppresses high-frequency signals. In contrast, wave propagation distributes energy more evenly across frequencies via oscillation, preserving both global coherence and fine detail. This makes wave-based propagation a strong complement to heat-based models with enhanced ability to capture complex patterns.


\begin{table*}[h]
\centering
\vspace{-5pt}
\setlength{\tabcolsep}{6pt}
\renewcommand{\arraystretch}{1.2}
\begin{tabular}{c|cccc|c}
\toprule
\textbf{Method} & \textbf{Image size} & \textbf{\#Param.} & \textbf{FLOPs} & \makecell{\textbf{Test Throughput} \\ \textbf{(img/s)}} & \makecell{\textbf{ImageNet} \\ \textbf{top-1 acc. (\%)}} \\
\midrule
$\text{Swin-T}_{\textcolor{blue}{\text{ICCV2021}}}$ & $224^2$ & 28M & 4.6G & 1242 & 81.3 \\
$\text{ConvNeXt-T}_{\textcolor{blue}{\text{CVPR2022}}}$ & $224^2$ & 29M & 4.5G & 1198 & 82.1 \\
$\text{DCFormer-SW-T}_{\textcolor{blue}{\text{WACV2023}}}$ & $512^2$ & 28M & 4.5G & - & 82.1 \\
$\text{Vim-S}_{\textcolor{blue}{\text{ICML2024}}}$ & $224^2$ & \textbf{26M} & 5.3G & 811 & 81.4 \\
$\text{vHeat-T}_{\textcolor{blue}{\text{CVPR2025}}}$ & $224^2$ & 29M & 4.6G & 1514 & 82.2 \\
\rowcolor{gray!20} WaveFormer-T (Ours) & $224^2$ & 29M & \textbf{4.4G} & \textbf{1560} & \textbf{82.5} \\
\midrule
$\text{Swin-S}_{\textcolor{blue}{\text{ICCV2021}}}$ & $224^2$ & 50M & 8.7G & 720 & 83.0 \\
$\text{ConvNeXt-S}_{\textcolor{blue}{\text{CVPR2022}}}$ & $224^2$ & 50M & 8.7G & 687 & 83.1 \\
$\text{DCFormer-SW-S}_{\textcolor{blue}{\text{WACV2023}}}$ & $512^2$ & 50M & 8.7G & - & 82.9 \\
$\text{vHeat-S}_{\textcolor{blue}{\text{CVPR2025}}}$ & $224^2$ & 50M & 8.5G & 945 & 83.6 \\
\rowcolor{gray!20} WaveFormer-S (Ours) & $224^2$ & 50M & \textbf{7.8G} & \textbf{1020} & \textbf{83.9} \\
\midrule
$\text{Swin-B}_{\textcolor{blue}{\text{ICCV2021}}}$ & $224^2$ & 88M & 15.4G & 456 & 83.5 \\
$\text{ConvNeXt-B}_{\textcolor{blue}{\text{CVPR2022}}}$ & $224^2$ & 89M & 15.4G & 439 & 83.8 \\
$\text{RepLKNet-31B}_{\textcolor{blue}{\text{CVPR2022}}}$& $224^2$ & 79M & 15.3G & - & 83.5 \\
$\text{DCFormer-SW-B}_{\textcolor{blue}{\text{WACV2023}}}$ & $512^2$ & 88M & 15.4G & - & 83.5 \\
$\text{Vim-B}_{\textcolor{blue}{\text{ICML2024}}}$ & $224^2$ & 98M & 19.0G & 294 & 83.2 \\
$\text{vHeat-B}_{\textcolor{blue}{\text{CVPR2025}}}$ & $224^2$ & \textbf{68M} & 11.2G & 661 & 84.0 \\
\rowcolor{gray!20} WaveFormer-B (Ours) & $224^2$ & \textbf{68M} & \textbf{10.8G} & \textbf{719} & \textbf{84.2} \\
\bottomrule
\end{tabular}
\caption{Performance comparison of image classification on ImageNet-1K.}
\label{tab:imagenet}
\vspace{-10pt}
\end{table*}

\section{Experiment}
\subsection{Experimental Settings}

We evaluate WaveFormer on three core vision tasks—image classification, object detection, and semantic segmentation—using ImageNet-1K~\cite{deng2009imagenet}, COCO~\cite{lin2014microsoft}, and ADE20K~\cite{zhou2017scene}. We use the same training strategy for each task to ensure a fair comparison. All models are trained using Pytorch, and some pre-trained weights are used for downstream tasks. More details are provided in Appendix 2.

\subsection{Experimental Results}

\paragraph{Image classification.} 
The ImageNet-1K classification results are summarized in Table~\ref{tab:imagenet}. Under similar FLOPs and model sizes, WaveFormer outperforms representative CNN and ViT architectures across all scales. For example, WaveFormer-T achieves 82.5\% top-1 accuracy, surpassing Swin-T/ConvNeXt-T by +1.2\%/+0.4\% with only 4.4G FLOPs. Notably, WaveFormer also exhibits clear advantages at larger scales. Specifically, WaveFormer-B reaches 84.2\% top-1 accuracy with 10.8G FLOPs and 68M parameters, improving over Swin-B/Vim-B by 0.7\%/1.0\% while being more computationally efficient. In terms of throughput, WaveFormer shows consistently higher inference speed than baselines: for instance, WaveFormer-T achieves 1560 images/s, which is 26\% faster than ConvNeXt-T and 92\% faster than Vim-S, while maintaining superior accuracy. These results highlight the efficacy of frequency-aware wave propagation for balanced global and local feature modeling.

\paragraph{Object Detection and Instance Segmentation.}
We further evaluate WaveFormer as a backbone for object detection and instance segmentation on the MS COCO 2017 dataset using Mask R-CNN. Classification-pretrained weights are used for fair downstream comparisons, with appropriate alignment for input resolutions. Results under both 1$\times$ and 3$\times$ training schedules are summarized in Table~\ref{tab:coco}. WaveFormer consistently outperforms Swin and ConvNeXt in both box AP ($\text{AP}^b$) and mask AP ($\text{AP}^m$), while maintaining comparable FLOPs. For example, with the 1$\times$ schedule, WaveFormer-T achieves 45.8\% AP$^b$ and 41.5\% AP$^m$, surpassing Swin-T by +3.1\%/+2.2\% and ConvNeXt-T by +1.6\%/+1.4\%. Similar trends hold for larger backbones: WaveFormer-B achieves 47.9\% AP$^b$ and 43.2\% AP$^m$, outperforming Swin-B and ConvNeXt-B. Moreover, WaveFormer consistently delivers higher inference FPS; for instance, WaveFormer-B runs at 20.4 images/s, which is 48\%/45\% faster than Swin-B/ConvNeXt-B. These improvements demonstrate the benefit of oscillatory bidirectional propagation for dense prediction tasks.

\paragraph{Semantic Segmentation.} 
We also benchmark WaveFormer on semantic segmentation using UperNet on the ADE20K dataset, and the results are shown in Table~\ref{tab:ade20k}. Across all scales, WaveFormer consistently achieves higher mean IoU (mIoU) while reducing computational cost. For example, WaveFormer-T obtains 47.4\% mIoU with slightly fewer FLOPs than ConvNeXt-T, improving over Swin-T by +3.0\%. At the base scale, WaveFormer-B reaches 50.5\% mIoU, outperforming NAT-B and ViL-B by 2.0\% and 1.7\%. Similar to detection results, WaveFormer maintains competitive or higher inference FPS compared to existing backbones. These results further validate that frequency-aware wave propagation preserves both global structure and fine-grained details, leading to superior semantic segmentation performance. Detailed case study segmentation results are provided in Appendix 5.

Overall, across classification, detection, and segmentation, WaveFormer consistently achieves a better accuracy–efficiency trade-off compared to both CNN- and ViT-based models. Its ability to balance low-frequency global coherence and high-frequency detail preservation proves crucial for complex visual tasks.

\begin{table}[ht]
\centering
\setlength{\tabcolsep}{5pt} 
\renewcommand{\arraystretch}{1.2} 
\begin{tabular}{c|c|c|c|c}
\toprule
\multicolumn{5}{c}{\textbf{Mask R-CNN 1$\times$ schedule on COCO}} \\
\midrule
\textbf{Backbone} & \textbf{AP$^b$} & \textbf{AP$^m$} & \makecell{\textbf{FPS}\\(img/s)} & \textbf{FLOPs} \\
\midrule
Swin-T      & 42.7 & 39.3 & 26.3 & 267G \\
ConvNeXt-T  & 44.2 & 40.1 & 29.3 & \textbf{262G} \\
vHeat-T & 45.1 & 41.2 & \textbf{32.7} & 272G \\
\rowcolor{gray!20} WaveFormer-T (Ours) & \textbf{45.8} & \textbf{41.5} & 32.1 & 270G \\
\midrule
Swin-S      & 44.8 & 40.9 & 19.7 & 359G \\
ConvNeXt-S  & 45.4 & 41.8 & 20.2 & 349G \\
vHeat-S & 46.8 & 42.3 & 25.9 & 348G \\
\rowcolor{gray!20} WaveFormer-S (Ours) & \textbf{47.0} & \textbf{42.5} & \textbf{26.2} & \textbf{345G} \\
\midrule
Swin-B      & 46.9 & 42.3 & 13.8 & 504G \\
ConvNeXt-B  & 47.0 & 42.7 & 14.1 & 486G \\
vHeat-B & 47.7 & 43.0 & 20.2 & 432G \\
\rowcolor{gray!20} WaveFormer-B (Ours) & \textbf{47.9} & \textbf{43.2} & \textbf{20.4} & \textbf{431G} \\
\midrule
\multicolumn{5}{c}{\textbf{Mask R-CNN 3$\times$ MS schedule on COCO}} \\
\midrule
Swin-T      & 46.0 & 41.6 & 26.3 & 267G \\
ConvNeXt-T  & 46.2 & 41.7 & 29.3 & \textbf{262G} \\
vHeat-T & 47.2 & 42.4 & 32.7 & 272G \\
\rowcolor{gray!20} WaveFormer-T (Ours) & \textbf{47.4} & \textbf{42.6} & \textbf{33.0} & 270G \\
\midrule
Swin-S      & 48.2 & 43.2 & 19.7 & 359G \\
ConvNeXt-S  & 47.9 & 42.9 & 20.2 & 349G \\
vHeat-S & 48.8 & 43.7 & 25.9 & 348G \\
\rowcolor{gray!20} WaveFormer-S (Ours) & \textbf{49.0} & \textbf{43.9} & \textbf{26.3} & \textbf{345G} \\
\bottomrule
\end{tabular}
\caption{Comparison of different backbones on COCO using Mask R-CNN with 1$\times$ and 3$\times$ MS training schedules. Best results in each group are highlighted.}
\vspace{-3ex}
\label{tab:coco}
\end{table}

\begin{table}[ht]
\centering
\setlength{\tabcolsep}{6pt}
\renewcommand{\arraystretch}{1.2}
\begin{tabular}{c|c|c|c}
\toprule
\multicolumn{4}{c}{\textbf{UperNet on ADE20K}} \\
\midrule
\textbf{Backbone} & \textbf{mIoU} & \makecell{\textbf{FPS}\\(img/s)} & \textbf{FLOPs} \\
\midrule
Swin-T & 44.4 & 31.8 & 237G \\
ConvNeXt-T & 46.0 & \textbf{37.8} & \textbf{235G} \\
ViL-S & 46.3 & - & - \\
vHeat-T & 46.9 & 36.7 & \textbf{235G} \\
\rowcolor{gray!20} WaveFormer-T (Ours) & \textbf{47.4} & 36.9 & \textbf{233G} \\
\midrule
Swin-S & 47.6 & 22.1 & 261G \\
NAT-S & 48.0 & 23.1 & 254G \\
ConvNeXt-S & 48.7 & \textbf{27.7} & 257G \\
vHeat-S & 49.1 & 26.1 & 254G \\
\rowcolor{gray!20} WaveFormer-S (Ours) & \textbf{49.8} & 26.4 & \textbf{252G} \\
\midrule
Swin-B & 48.1 & 19.2 & 299G \\
NAT-B & 48.5 & 20.8 & \textbf{285G} \\
ViL-B & 48.8 & - & - \\
ConvNeXt-B & 49.1 & 21.6 & 293G \\
vHeat-B & 49.6 & 23.6 & 293G \\
\rowcolor{gray!20} WaveFormer-B (Ours) & \textbf{50.5} & \textbf{23.8} & 290G \\
\bottomrule
\end{tabular}
\caption{Comparison of different backbones on ADE20K using UperNet. Best results in each group are highlighted.}
\vspace{-2ex}
\label{tab:ade20k}
\end{table}

\begin{figure}[ht]
    \centering
    \includegraphics[width=\linewidth]{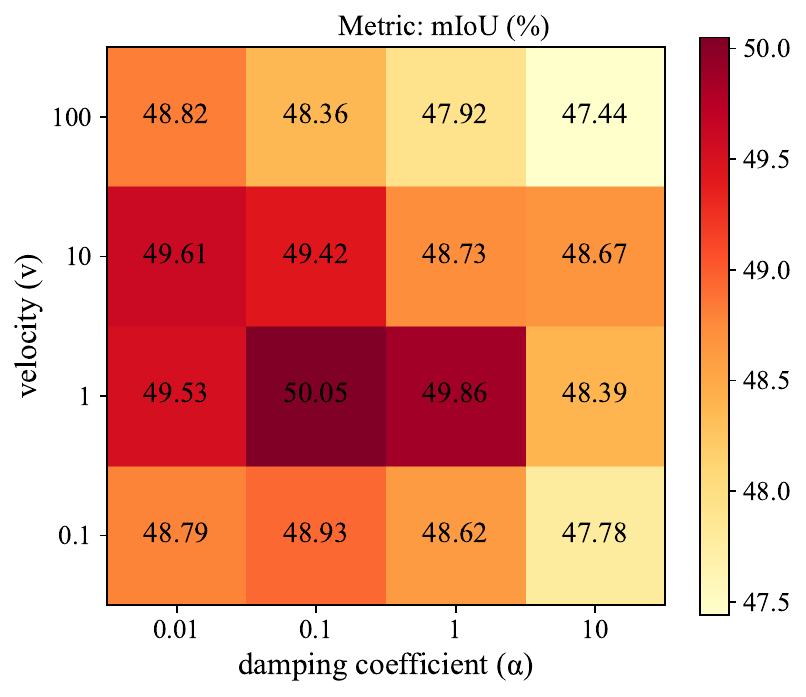} 
    \caption{Evaluation of thermal wave velocity and damping term using WaveFormer-B with different $(v, \alpha)$ settings.}
    \label{fig:pic/ablation}
\end{figure}

\subsection{Experimental Analysis}
\paragraph{Damping-Controlled Spectral Modulation Capability.}

To investigate the role of wave dynamics in semantic propagation, we conduct an ablation study on WaveFormer-B by varying the wave velocity $v \in {0.1, 1, 10, 100}$ and damping coefficient $\alpha \in {0.01, 0.1, 1, 10}$. As shown in Figure~\ref{fig:pic/ablation}, the best performance is achieved at $v = 1$, $\alpha = 0.1$, indicating that moderately damped oscillations offer the most effective trade-off. Extreme values of either parameter degrade results—too little damping leads to instability, while excessive damping suppresses high frequencies. This trend is supported by the spectral condition in Eq.~\eqref{9}, where the effective oscillation frequency $\omega_d$ exists only if $v^2(\omega_x^2 + \omega_y^2) > (\alpha/2)^2$, suggesting that strong damping or low wave velocity reduces the available frequency bandwidth. Thus, $v$ determines how broadly semantic signals propagate across frequencies, while $\alpha$ controls their persistence. The optimal region balances temporal coherence and spectral richness, enabling precise, frequency-aware modulation for smooth yet detailed semantic representation.

\paragraph{Frequency-Time Decoupled Semantic Propagation for Fine-Grained Boundary Modeling.}
WaveFormer performs comparably to vHeat in global tasks, but surpasses it in segmentation with fine structural demands. This stems from their spectral behavior: vHeat’s heat kernel $e^{-k \omega^2 t}$ strongly attenuates high frequencies, leading to oversmoothing.  In contrast, WaveFormer adopts frequency–time decoupled wave dynamics (Eq.~\ref{7}), where high-frequency components are preserved through oscillatory terms $\cos(\omega_d t)$ and $\sin(\omega_d t)$, while the uniform damping term $e^{-\alpha t/2}$ provides controllable temporal decay, while maintaining detail and boundary sharpness. Segmentation case studies (Fig.\ref{fig:pic/case}) confirm WaveFormer consistently yields sharper boundaries. Its oscillation-driven propagation avoids the blurring seen in heat-based methods, highlighting the benefits of frequency-time decoupling for fine-grained boundary modeling.

\section{Conclusion}
We propose WaveFormer, a physics-inspired vision backbone that models semantic propagation through the damped wave equation. By leveraging the frequency–time decoupling of the wave equation, our formulation enables oscillatory information flow that preserves both global context and fine-grained structure. To instantiate this, we develop the Wave Propagation Operator (WPO), an efficient and interpretable module that simulates global semantic communication with $\mathcal{O}(N \log N)$ complexity. WaveFormer achieves strong performance across classification, detection, and segmentation tasks, demonstrating the effectiveness of wave-based modeling.


\appendix

\clearpage

\section{Acknowledgments}
This work was supported in part by Natural Science Foundation of China (No. U24B6012, 61972217, 32071459, 62176249, 62006133, 62271465, 62406167, 12125401), Shenzhen Medical Research Funds in China (No. B2302037), National Key Research and Development Program of China (No. 2023YFF1204400 and 2023YFF1204401), and AI for Science (AI4S)-Preferred Program, Peking University Shenzhen Graduate School, China.

\bibliography{aaai2026}

\end{document}